\documentclass[letterpaper, 10 pt, conference]{ieeeconf}  

\IEEEoverridecommandlockouts                              

\usepackage{graphicx} 
\usepackage{booktabs}
\usepackage{cleveref}

\title{\LARGE \bf
Inconsistency-based Active Learning for LiDAR Object Detection
}

\author{Esteban Rivera$^{1}$, Loic Stratil $^{1}$ and Markus Lienkamp $^{1}$
\thanks{This project has beeing developed in the scope of the BFS-DAD (Bayerische Forschungsstiftung - Data-based Autonomous Driving) project}
\thanks{$^{1}$Institute for Automotive Engineering,
        Munich Institute of Robotics and Machine Intelligence
        Technical University of Munich,
        {\tt\small name.lastname@tum}}%
}

\begin{document}

\maketitle
\thispagestyle{empty}
\pagestyle{empty}

\begin{abstract}

Deep learning models for object detection in autonomous driving have recently achieved impressive performance gains and are already being deployed in vehicles worldwide. However, current models require increasingly large datasets for training. Acquiring and labeling such data is costly, necessitating the development of new strategies to optimize this process. Active learning is a promising approach that has been extensively researched in the image domain. In our work, we extend this concept to the LiDAR domain by developing several inconsistency-based sample selection strategies and evaluate their effectiveness in various settings. Our results show that using a naive inconsistency approach based on the number of detected boxes, we achieve the same mAP as the random sampling strategy with 50\% of the labeled data.

\end{abstract}

\section{Introduction}
\label{sec:intro}
In recent years, several companies have deployed autonomous fleets globally\cite{waymo_blog}. Despite this progress, the widespread adoption of autonomous driving  remains a challenge. From a perceptual standpoint, one of the primary challenges is the need for detection models to process vast amounts of sensor data originating from diverse environments and scenarios encountered by autonomous vehicles. The cost of capturing, processing, storing, and labeling this data can become increasingly expensive, even for the industry's leaders. Consequently, a major focus within both the industry and academic research is to either reduce reliance on labeled data or optimize the usage of labeling budget to get the maximal performance gains per dollar spent.

Active Learning (AL) is a strategic approach designed to address data scarcity and constraints on labeling budgets, not only for computer vision models but for the general domains. It leverages deep learning models applied to real-world tasks, focusing on the principle that models trained for specific tasks can also identify the most valuable unlabeled samples for labeling \cite{ren2022}. Instead of randomly labeling data, this method prioritizes samples that promise the most significant informational gain or improvement potential. The underlying hypothesis is that this targeted labeling strategy reduces the amount of data needed, and consequently, the budget required to achieve acceptable performance levels on a given task.
\begin{figure}[ht]
  \centering
  \includegraphics[width=\linewidth]{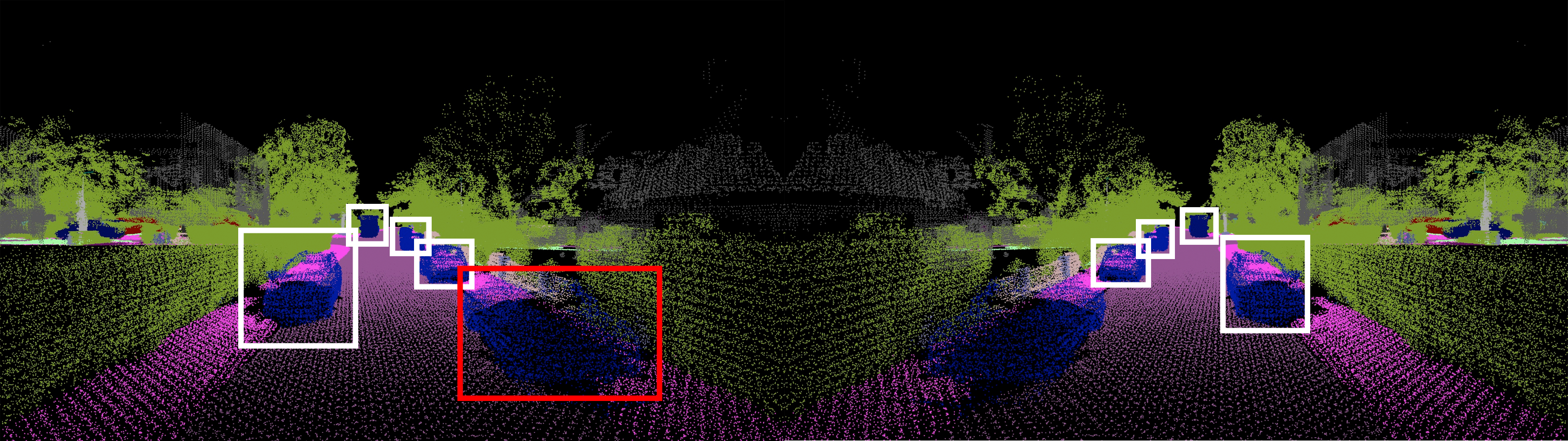}
  \caption{Point cloud and its augmented version side to side, with white colored boxes representing objects which where consistently recognised with a 3d Detector in both point clouds, and a red box representing one car which was recognised only in one point cloud. This is an example of a point cloud which is interesting to label, because the model lacks robustness in its representation. Image from the KITTI dataset \cite{Geiger2012CVPR}.
  }
  \label{fig:intro}
\end{figure}
AL has been extensively applied in the domain of computer vision with notable success, as evidenced by various studies \cite{Haussmann2020,ren2022}. Similarly, inconsistency approaches for AL, which assess discordance between model predictions to select data points for labeling, have also demonstrated their effectiveness\cite{elezi2022,sohn2020,gao2020,ghita2024activeanno3d}. In this paper, we aim to explore the potential of these methodologies within the LiDAR domain for 3D object detection. To the best of our knowledge, this application has not been extensively investigated previously.

Several methods exist that leverage trained detector models to identify the best samples for labeling. Approaches focusing on uncertainty and entropy are well-documented in the literature \cite{wu2022}. Additionally, diversity-based methods have also shown promising results \cite{Liang2022ExploringDA}. Another notable strategy is the ensemble, or query-by-committee approach\cite{seung1992}, which serves as a proxy for sample uncertainty. This method considers a sample to be uncertain if multiple models produce contradictory predictions about it. Such discrepancies may indicate that the sample deviates from the distribution used in training, or lies close to the decision boundary, making classification challenging. Given the variety of models and datasets available today, this strategy has become increasingly popular in recent years.

Inconsistency can be evaluated following two strategies: On one hand multiple models can be utilized to compare their outputs to identical inputs. On the other hand only one model can be used by applying certain types of augmentations where the output result remains unchanged, or the change can be replicated. Based on the latter strategy we propose using a simple mirroring augmentation following \cite{elezi2022} specifically for the LiDAR domain. With that, we predict the bounding boxes for the original and the mirrored point clouds and then compare them through several scoring strategies to quantify the level of inconsistency between the two predictions, see \Cref{fig:intro}. To quantify inconsistency we propose two methods with increasing complexity, which quantify the level of per-sample information for the selection algorithm for active learning. We then use the selected samples with different training paradigms (Fine-tuning, Retraining) to further characterize which scenarios, training strategies and inconsistencies result in larger improvements.

Our contributions can be summarized as follows:
\begin{itemize}
    \item We developed a LiDAR-only sample selection strategy based on augmentation inconsistency.
    \item We improved the random baseline mAP up to 2.5\% for the complete data range.
    \item We designed, analysed and evaluated several strategies to sample point clouds based on inconsistency scores.
\end{itemize}

\section{Related work}
\subsection{Active Learning}
Active Learning is one of the most popular strategies to optimize the time and monetary budget for labeling data. Even though its concept is long established\cite{seung1992}, recent advances are traceable to the deep learning explosion. Several variations and techniques \cite{ren2022} have been developed for all the possible data domains. In particular, for computer vision, active learning approaches have been applied to improve efficiency for classification tasks \cite{Caramalau_2022_BMVC,Ranganathan2017}. Specifically, for 2D detection tasks, the approaches combine both localization and classification to improve the informativeness score of the selected samples. For example, Choi\cite{choi2021active} proposes using mixture density networks to estimate a probabilistic distribution for each localization and classification head's output. Aithal \cite{aithal23} utilizes an uncertainty measure that considers detection, classification, and distribution statistics to sample images with higher object prediction scores, aiming for a balanced distribution in sampling. To take advantage of all the possible predicted bounding boxes of a sample, Wu\cite{wu2022} proposes an entropy-based-non-max-supression assigning an uncertainty score to each box instead of the whole image, discarding the samples with the most redundant information. Finally, Haussman\cite{Haussmann2020} highlights the details and challenges of the implementation of an active learning pipeline for a real use case, namely, autonomous driving.

\subsection{Inconsistency analysis}
Inconsistency strategies aim to identify samples from the contradictory outputs of multiple models, or multiple views from the same scene. Alternatively, one model predicts inconsistent outputs when queried with augmented inputs whose output should remain unchanged. Seung \cite{seung1992} proposed the \textit{Query by Committee} approach, where samples for training a simple perceptron are identified from the maximum disagreement between pretrained models. They also studied the theoretical information gain obtained by this approach compared to random sampling. Specifically, in the domain of computer vision, several approaches have used the consistency approach to train models in a semi-supervised manner \cite{sohn2020,gao2020}, where consistency between augmented samples is enforced during training. Later, further optimization of the training budget was achieved by simultaneous application  of the inconsistency approach to active and semi-supervised learning strategies \cite{elezi2022,lim2023active,cai2022,rangnekar2022,chen2023,huang2021}. Thereby, the inconsistency is used by the active learning approach to find the best samples to label and by the semi-supervised approach to ensure robustness against augmentations.

\subsection{3D active learning}
Active learning has been applied already within the domain of 3D object detection in the context of autonomous driving, a key area of computer vision. Here, both camera images and LiDAR point clouds - which are more capable of capturing the 3D nature of the environment - are leveraged. Schmidt \cite{schmidt2020} compares the performance of several 2D and 3D active learning approaches for camera object detection. Kao \cite{kao2018} explores how the localization information of the objects influences the quality of the selected samples.

The following works take advantage of the broader sensor availability of the autonomous driving application instead of using only camera data. For instance, Liang \cite{Liang2022ExploringDA} uses information from GPS data to select samples recorded in different map positions, ensuring they represent a variety of scenarios. Furthermore, several works attempt to use information distillation between camera and LiDAR to select the best samples to label. Hekimoglu \cite{Hekimoglu_2024_WACV} uses the 3D information of the LiDAR point cloud to find the most informative samples for camera training. On the other hand, Gunnard \cite{gunnard2021} proposes improving the sample selection for both LiDAR and camera models in parallel. Finally, Rivera \cite{rivera2024} proposes comparing the inconsistencies between camera and LiDAR in the 2D ego plane to compensate for the lack of depth estimation in camera detectors and find informative samples.

\section{Method}
The core of our method is to develop a new strategy to sample LiDAR point clouds in an Active Learning setting. We therefore define here an active learning cycle and the concept of inconsistency to quantify the usefulness of a given sample.
\subsection{Active Learning cycle}
In this paper we present an active learning cycle as follows: Using the KITTI dataset, we begin by training a 3D detection model with 10\% of the available labeled data, namely 371 point clouds. This serves as starting point for the Active Learning cycle. Firstly, the model is used to get the bounding boxes predictions on the remaining 90\% of the active learning dataset, or working set. Secondly, we select samples in accordance with the developed strategy, amounting to 10\% of the original dataset per cycle, namely 371 samples from the working set. Thereby, the ground truth label is used to simulate a human labelling process. Thirdly, these 371 samples are then added to the labeled training set and the object detector is trained with it. This is a complete active learning cycle. The cycle is repeated until the available labeled data has been completely used, adding 10\% of the original dataset on each iteration. The random baseline, used for a quantitative comparison of the different sample selection approaches leverages random sample selection in every cycle and is trained from scratch in each iteration.

\subsection{Inconsistency definition}
To select the interesting samples for the active learning cycle, an inconsistency-based approach is proposed. The approach is based on the Query-by-committee idea, where instead of using several models to get the predictions on a single sample, we propose augmenting the sample, in our case a LiDAR point-cloud, and get the predictions for both the original and the augmented samples. The main idea behind this, similar to Elezis work \cite{elezi2022}, is that the model should -in general- be robust to weak augmentations, like mirroring, or small shifts on the input. Therefore, samples for which the model is not robust and presents inconsistent detections should be considered for ground truth labeling. Compared to the pure 2D image case, the diversity of point-cloud augmentations is limited, because brightness, contrast, or colorization augmentations are no longer available. Therefore, we focus only on a horizontal reflection of the point cloud, because this augmentation represents a scenario which occurs in real life. 

The next aspect to consider is how to compare between the original and the augmented point cloud to find inconsistencies. Intersection over Union (IoU) has been already proposed in the literature for both camera 2D and 3D detections \cite{Hekimoglu_2024_WACV}; and class mismatches have been used for 2D detection \cite{elezi2022}. To that regard, we compare the number of boxes against each other\cite{rivera2024} for a single modality LiDAR approach. From this comparison, an inconsistency score $S_{NoB}$ is:
\begin{equation}
    S_{NoB} = \frac{|N_o - N_a|}{max(N_o, N_a)},
  \label{eq:important}
\end{equation}
where $N_o$ is the number of boxes detected on the original point cloud and $N_a$ is the number of boxes detected on the augmented point cloud. The score is normalised over the maximal number of boxes detected, to have a notion of relative inconsistency. We discard samples for which both $N_o$ and $N_a$ are 0. The relative inconsistency metric is motivated with the following example: A sample for which two boxes are detected with four detections in the augmented version is - relatively - more inconsistent than a sample where 20 boxes are detected and 22 boxes are detected in the augmented version. Even though the absolute bounding box inconsistency amounts to two in both cases, the relative inconsistency is 0.5 for the first case and 0.09 for the second. We call this approach \textit{Number-of-boxes inconsistency score (NoB)}.

For the sake of completeness, we also evaluate the performance of an IoU-based inconsistency score, leveraging 3D IoU for cross-modal comparison. Only if this value is above a threshold we accept the box as a match. The number of matches is summed as $N_m$ and then the score $S_{IoU}$ is calculated as:
\begin{equation}
    S_{IoU} = \frac{max(N_o, N_a)-N_{m}}{max(N_o, N_a)} .
\label{eq:edge2}
\end{equation}

\section{Experiment}
\label{sec:experimentdefinition}
Besides testing the different inconsistency calculation methods, we tested two strategies to use the selected inconsistency in the training of the model. On the one hand, in the \textbf{Scratch} setting, we trained the model from scratch with the samples selected within the current cycle in addition to the samples selected in previous cycles. On the other hand, in the \textbf{Retrain} setting, we took the trained checkpoint from the previous iteration as starting point and trained it further for 80 epochs for each active learning cycle. We tested the \textbf{Retrain} setting for both the $S_{NoB}$ and the $S_{IoU}$ inconsistency types.
\subsection{Model}
All the experiments were conducted with the PointPillars architecture due to its training speed. Given that we do not aim for the best performance, but rather look for the greatest relative improvement, a fast model allowed us to perform more experiments in a limited time window. The training settings were kept constant across the experiments and defined as:
Epochs: 80, learning rate: 0.001, Optimizer: ADAM. The KITTI dataset was used for the experiments. Only the labeled original train set with 7481 point clouds was used. It was split into two subsets: 3712 for the Active learning cycle and 3769 for test.

\section{Results}
\subsection{Inconsistency ordering}
Using a quantitative inconsistency metric for sample selection allows to order them in descending or ascending order. Thus, as a proof of concept, we design an experiment to preliminary compare the two possible orderings, either descending or ascending. In both orderings, the samples with inconsistency greater than 0 are used first. When there are no more of such samples, the remaining samples are used to complete the training. 

\begin{figure}[tb]
  \centering
  \includegraphics[height=6.5cm]{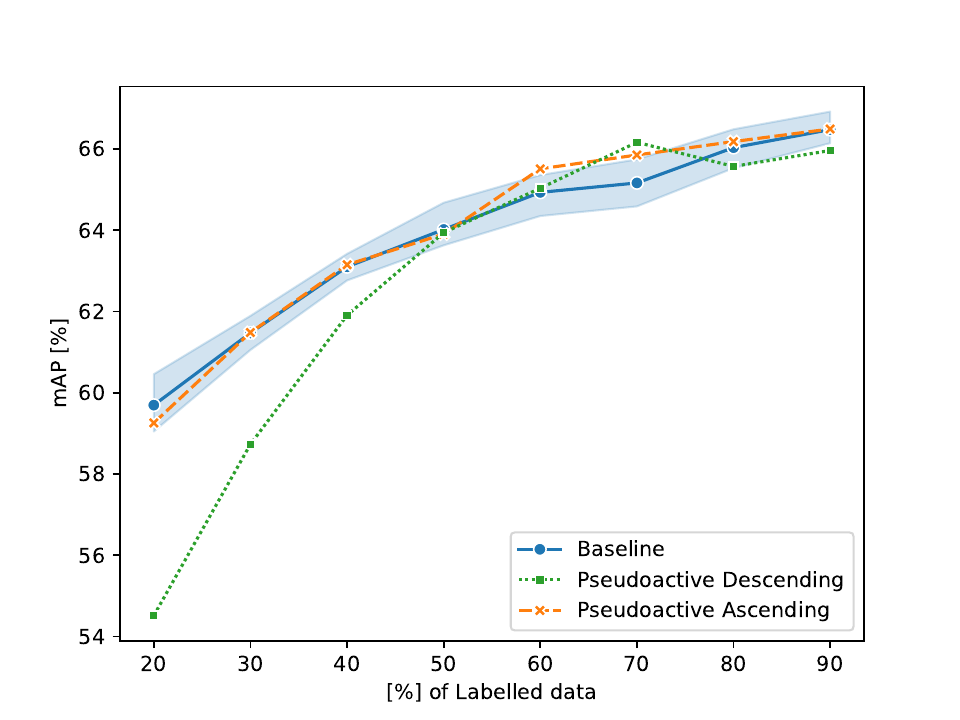}
  \caption{Pseudo active learning cycle results. Deviation for the baseline is shown as the blue area
  }
  \label{fig:pseudoexperiment}
\end{figure}

We evaluate the effectiveness of the ascending vs descending sampling strategy with a pseudo-active learning cycle. Its function is analogous to the active learning cycle with the key difference being that samples are ordered once only during the first cycle. This sample order is then kept until all samples are selected. Thus, in each subsequent cycle the most suitable 10\%, i.e. the next chunk of samples, is added to the training set until all the samples are used. The performance results are shown in  \Cref{fig:pseudoexperiment}. For the random baseline the standard deviation is plotted as a region as well.

Contrary to our expectations, leveraging samples in descending order performs worse than the random baseline for the low-data regime (20\%-40\%). When using more than 50\% of the available data it has a similar performance as the random baseline. For the ascending approach, there is no clear improvement in comparison to the random sampling, but is stable across the whole data range. In accordance with these findings, we utilize an ascending ordering to select our samples for the subsequent active learning cycle.

\subsection{Active learning cycle}
\begin{figure}[tb]
  \centering
  \includegraphics[height=6.5cm]{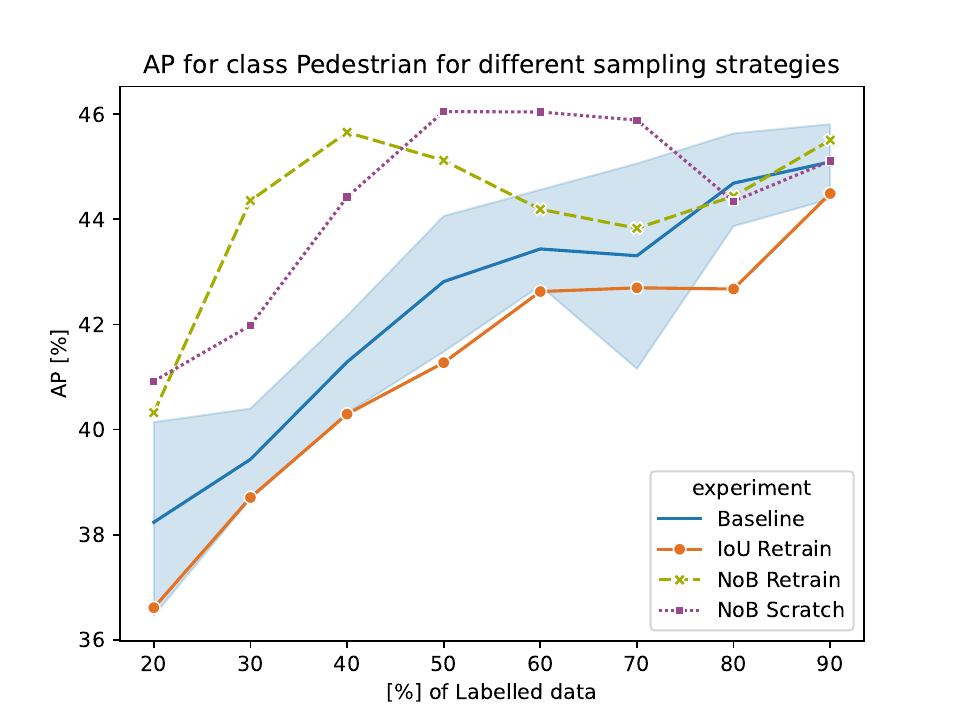}
  \caption{Results for Pedestrian
  }
  \label{fig:pedestrianactivel}
\end{figure}

\begin{figure}[tb]
  \centering
  \includegraphics[height=6.5cm]{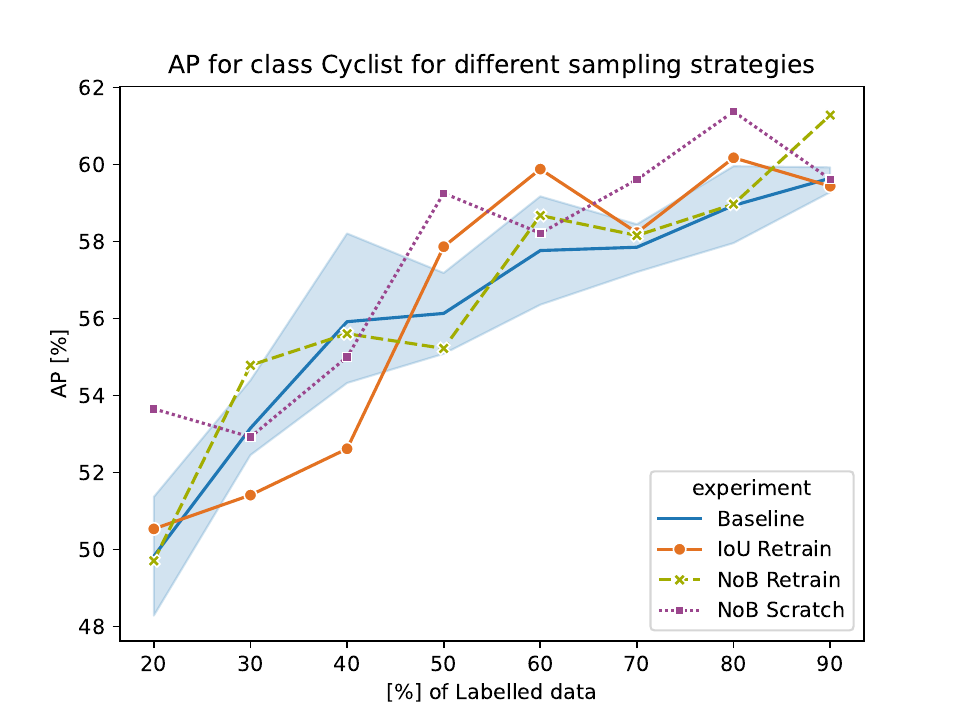}
  \caption{Results for Cyclist
  }
  \label{fig:cyclistal}
\end{figure}

\begin{figure}[tb]
  \centering
  \includegraphics[height=6.5cm]{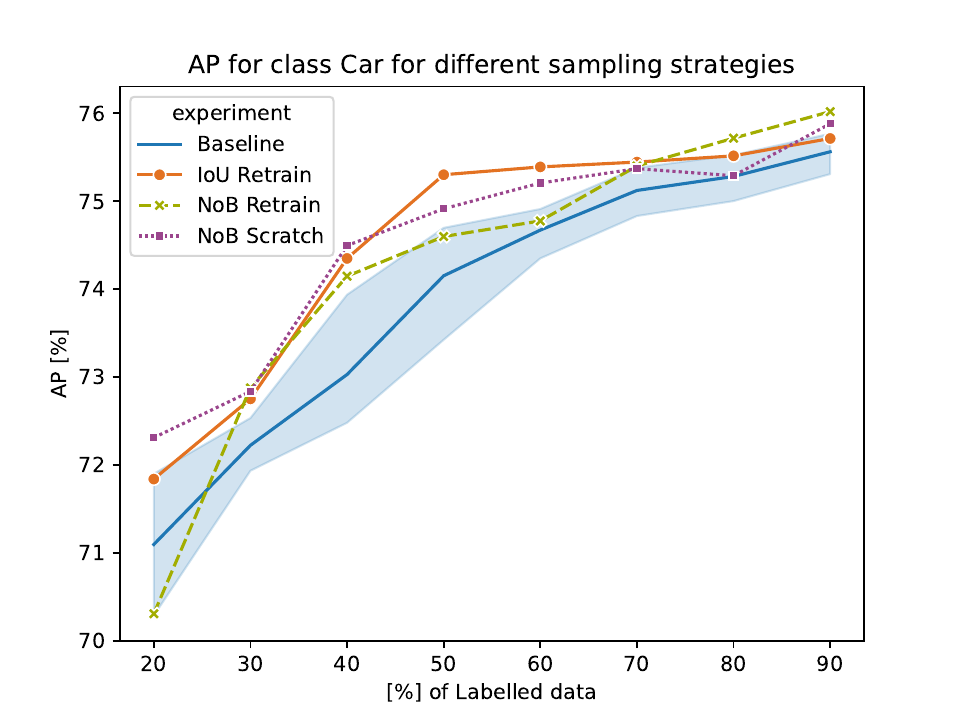}
  \caption{Results for Car
  }
  \label{fig:caral}
\end{figure}

\begin{figure}[tb]
  \centering
  \includegraphics[height=6.5cm]{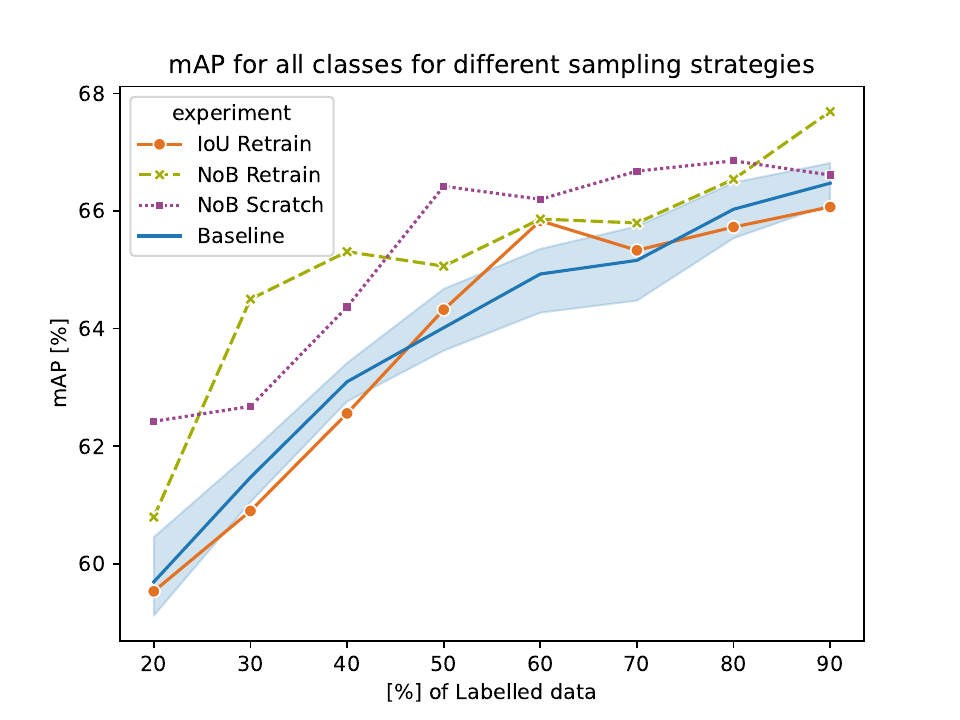}
  \caption{Results for mAP (all classes)
  }
  \label{fig:maxal}
  \vspace{-0.5cm}
\end{figure}

Results are presented class-wise, namely pedestrian, cyclist and car, because the AP curve along the range of labeled data shows different behaviors.
In each of the plots, the three settings explained in \Cref{sec:experimentdefinition} and the random baseline are shown. Each experiment was run three times with different seeds and the best value is plotted. 

For the pedestrian case in \Cref{fig:pedestrianactivel}, an interesting trend is seen for both the NoB Scratch and the NoB Retrain settings. On the low- to medium-data regime (20\%-70\%), they both present better AP than the baseline, showing a steep increment between 30 \% and 40\% of the available data. For the NoB Retrain setting, the AP reaches a peak at 40\% of labeled data before progressively descending to the baseline values. On the other side, the NoB Scratch setting, reaches a plateau until 70\% of the available data. Finally, the IoU setting underperforms along the entire labeled data range in comparison to the baseline.

The results for cyclist, seen in \Cref{fig:cyclistal}, are interesting from the labelling efficiency side, because the class is underrepresented in the whole KITTI dataset, with only about 4.67\% of the labels. Therefore, it is interesting to see how the proposed strategies behave when there are even fewer instances of the desired class. It can be seen that for the 40\% of the data or less, the three settings have lower or equal AP than the random baseline. For both the NoB Retrain  settings, this trend continues for the whole range. On the other hand, the NoB Scratch setting is slightly better than the random baseline, with 1-2\% on average.

Contrary to the cyclists, the car is the most represented class in the dataset with about 82.5\% of all labels. As seen in \Cref{fig:caral}, the behavior of all settings is more stable than before. In this case, all the three settings present a better AP than the baseline. The NoB Scratch and the IoU Retrain are stronger in the medium-to-low data regime, with AP at least 1\% above the baseline, while NoB Retrain performs slightly better on the high data regime. 

In the \Cref{fig:maxal} the mAP results for all the classes can be observed. The NoB Scratch setting yields the best results compared to the baseline, with an improvement between 1\% and 2\% across the range. From a monetary perspective, it achieves the same mAP as the random sampling with half of the labeled point clouds. The NoB Retrain setting is above the random baseline as well, with a more pronounced improvement on the low-data regime, where with only 30\% of the data achieves the same performance as the random baseline with 60\% or NoB Scratch with 40\%. On the other hand, the IoU-based setting underperfoms the random baseline across the majority of data range except between 50\% and 70\% of the data.

\subsection{Class proportion}
\begin{figure}[tb]
  \centering
  \includegraphics[height=6.5cm]{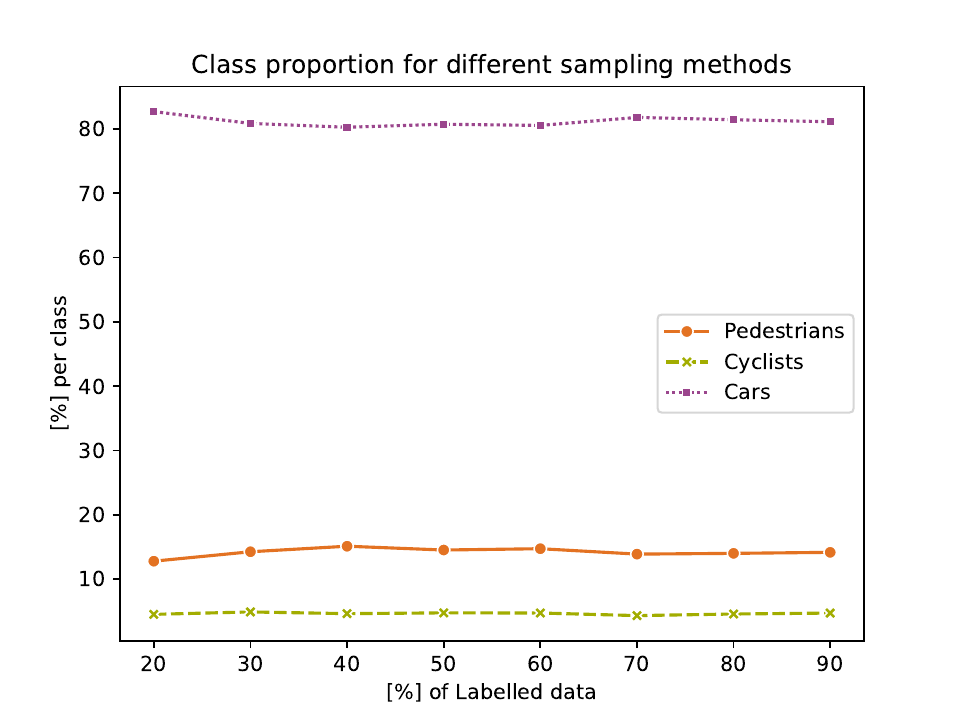}
  \caption{Class distribution across cycles
  }
  \label{fig:classdistribution}
  \vspace{-0.5cm}
\end{figure}


In \Cref{fig:pedestrianactivel,fig:cyclistal,fig:caral}, it can be observed that the behavior of all classes is not constant across the available data. For the pedestrian case \Cref{fig:pedestrianactivel}, the active learning approaches deliver better results than the baseline on the low to medium-data regime, reaching its maximum at the end of the range; for the cyclist, \Cref{fig:cyclistal}, even though the curve is not smooth, a small improvement can be seen in the high-data regime. Finally, for the car class \Cref{fig:caral}, there is a constant, but small improvement across the whole data range. To explain this behavior, we plotted the evolution of the class distribution across the data range for the NoB Scratch setting in \Cref{fig:classdistribution}. It can be seen that the class distribution does not vary along the data range. Therefore, the changes of the curves along the range are not caused by changes in the distribution of the data.

\subsection{Inconsistency proportion}
\begin{figure}[tb]
  \centering
  \includegraphics[height=6.5cm]{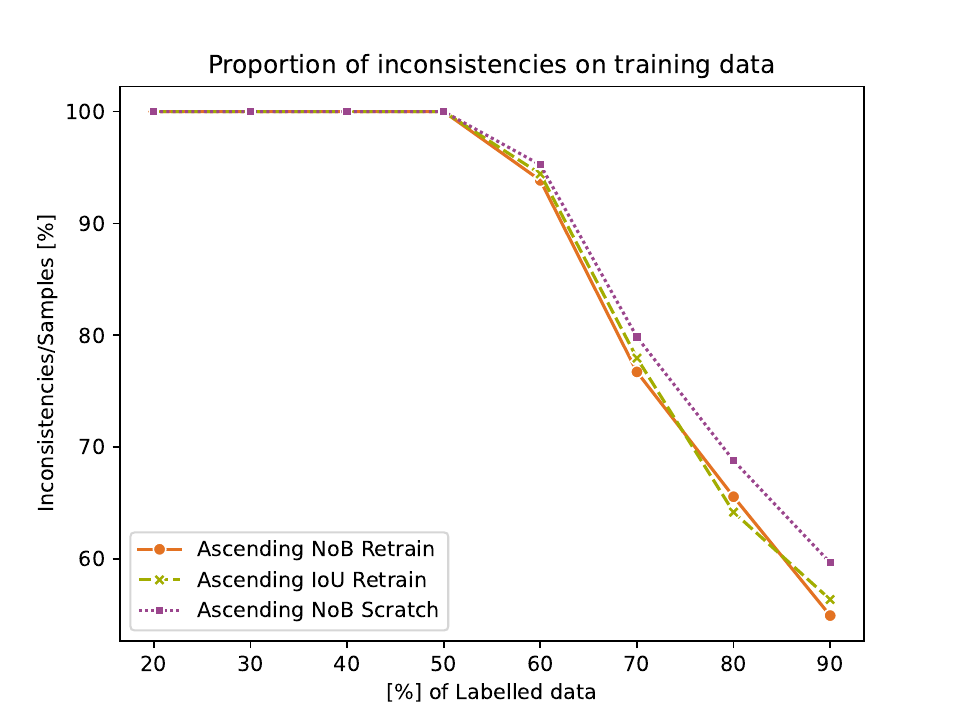}
  \caption{Inconsistency distribution across cycles
  }
  \label{fig:inconsistencyproportion}
  \vspace{-0.5cm}
\end{figure}

It can be seen from the results, that the relative performance with respect to the random baseline sampling is different for the low- medium- and high-data range. One of the possible causes might be due to the proportion of inconsistent samples found in each iteration. The larger the training set is compared to the total available data, the fewer inconsistent samples remain to be found. Therefore, with increasing active learning cycles, fewer inconsistent samples are found, even with the inclusion of consistent samples to meet the selection requirements per cycle, which in the end is the random sampling strategy. In the \Cref{fig:inconsistencyproportion} the inconsistency proportion for the three proposed experiments is shown. Noticeably, the behavior is similar for all experiment variations, with the active ascending approach having slightly more inconsistencies at the end of the range. As a main observation, the training set is composed of inconsistent samples exclusively up until 50\% of the available data. Above that distribution, consistent samples are selected for active learning cycles. Then, the models even after they are retrained within each iteration, do not find more inconsistent samples and start using directly the consistent ones.
\subsection{Further experiments}

\begin{table*}
\centering
\caption{Improvement of several settings with respect to the random baseline}
\label{tab:ablation}
\begin{tabular}{lcccr}
\toprule
Ordering & Inconsistency & Training & Normalised & Improvement (\%) \\
\midrule
Ascending & NoB & Retrain & Yes & 2.43 \\
Ascending & NoB & Scratch & Yes & 2.58 \\
Ascending & NoB & Fine-tuning & Yes & -5.76 \\
Ascending & IoU & Retrain & Yes & -0.13 \\
Ascending & NoB & Retrain & No & -0.11 \\
Descending & NoB & Retrain & Yes & -0.05 \\
\bottomrule
\end{tabular}
\vspace{-0.5cm}
\end{table*}

We additionally evaluated how the training type, ordering or inconsistency type would affect the performance of the sampling strategy, as seen in \Cref{tab:ablation}.
Related to the type of inconsistency, it can be seen that the naive NoB inconsistency outperforms the IoU inconsistency in the same retrain setting by almost 3\%, where the IoU inconsistency is actually worse than the random baseline. Using the NoB inconsistency, but fine tunning the model in each cycle only with the new inconsistencies is almost 8\% worse than the training with the new samples plus the ones from the previous iteration. The retrained case, where the checkpoint from the previous iteration is used as starting point, is 0.15\% worse than the best case, and therefore is still better as the random baseline. Finally, skipping the normalisation for the inconsistency score underperforms the random baseline by 0.11\%.

\section{Discussion}
Upon reviewing the presented results, several insights emerge. Firstly, examining \Cref{eq:important}, it becomes evident that samples with a higher number of detected objects exhibit lower inconsistency scores compared to those with fewer objects, which is attributable to the normalization of the score. Secondly, the analysis of both \Cref{fig:pseudoexperiment} and \Cref{tab:ablation} highlight a clear trend: The most effective sampling strategy is ascending, prioritizing samples with lower scores over those with higher ones. Consequently, these findings suggest that the optimal sampling strategy involves prioritizing samples with more objects during the first cycles, leveraging the inconsistency score as a proxy for object count. This approach proves particularly beneficial from a monetary standpoint in scenarios where labeling costs are incurred per frame rather than per bounding box.

A class-wise performance comparison leads to further interesting observations. With the "car" class, the improvement observed across the spectrum of labeled data, including the random baseline, remains relatively marginal, hovering around 1\%. This suggests that the performance for this class is approaching saturation, making even small improvements over the baseline promising. Moreover, the average precision (AP) of all sampling strategies remains constant and stable above the random baseline, suggesting that the observed improvement isn't just a result of random fluctuations within the dataset or the models themselves.

In comparison, the "cyclists" and "pedestrians" classes, experience steeper performance gains across the range of labeled data, with a difference of approximately 10\% AP between the initial and final data points. Notably, the AP improvement for "cyclists" is strongly fluctuating and gains are predominantly observed in the high-data range (70 - 90 \%). In contrast, the "pedestrians" class - with both number-of-box approaches - consistently outperform the baseline throughout the data range. Importantly, the "cyclists" class is strongly underrepresented from a class distribution standpoint - in comparison to the "cars" and even "pedestrians" class. This suggests the existence of a lower sample size limit, under which the available data may not provide sufficient information to allow for improvements using our sampling strategies.
The "pedestrians" class however, yields substantial improvements despite their under-representation relative to the "cars" class. 

This finding supports the following idea: For our active learning sampling strategies to work efficiently, a class must possess enough unused information, or entropy level, to allow for impactful training cycles of the detector. In our case, the "cars" class is already well-detected after the initial detector training due to the large "cars" sample set, leading to a low information gain. The "pedestrians" class is not robustly detected by the initial detector, however the available sample set is large enough to allow for impactful training, leading to a high information gain. The "cyclists" class, as stated above, is under-represented, resulting in a low information gain. 

Interestingly, it is insightful to examine the differences between the two types of inconsistencies under identical training methodologies, specifically concerning the number of boxes and the (IoU). As noted in several previous studies \cite{Hekimoglu_2024_WACV,schmidt2020,gunnard2021}, the inconsistency strategy for the 3D detection approach typically relies on IoU matching between bounding boxes across samples for comparison. Similar to \cite{rivera2024}, our results indicate that the consistency strategy based on the number-of-boxes surpass those based on IoU across all data ranges, achieving an average improvement of 2.56\%. We hypothesize that this superiority stems from the inherent limitations of IoU inconsistency, which is closely tied to the precise localization of boxes. For objects positioned at greater distances from LiDAR, IoU matching becomes highly susceptible to minor variations, rendering the inconsistencies derived from it excessively noisy and overshadowing any potentially useful information that could aid in training with localization error noise. On the other hand, using the number-of-boxes as a direct metric for inconsistency smoothens the score used to select the samples. Therefore, the strategy is more robust to small variations in the localisation and focusing it on finding the features which make the sample informative for the training.

Finally, we outline several directions for future work to further enhance the understanding in this field. Firstly, we plan to incorporate additional datasets. While the KITTI dataset remains highly regarded within the community, its relatively small size constrains the feasible depth of analysis, especially in terms of data budget constraints. Secondly, integrating a semi-supervised learning approach alongside active learning has been demonstrated to significantly enhance performance under limited data budgets \cite{elezi2022}. Furthermore, considering the relative success of the number-of-boxes inconsistency metric, it could be beneficial to directly incorporate this metric into the loss function. Such an integration would enforce consistency regarding the number of detected boxes during training, potentially leading to more robust models. Finally, to evaluate the transferability of our proposed method, it would be intriguing to adapt existing state-of-the-art active learning strategies for 3D detection, which currently rely on IoU-based inconsistency measures, to instead utilize inconsistency based on the number of boxes.

\section{Conclusion}
In this paper, we introduce a novel strategy for selecting LiDAR samples in an Active Learning framework, based on inconsistencies between a point cloud and its horizontally mirrored augmentation. We explore two consistency scores: one based on the number-of-boxes approach and another based on the IoU between matched boxes, comparing both to a random sampling baseline. Our analysis separates the results across the three KITTI dataset classes, offering insights into the behaviors observed. Additionally, we investigate the impact of the training approach and score normalization on our method's performance. Our findings indicate that a consistency score based solely on the number of boxes surpasses those based on IoU and the random baseline in our testing scenarios. This insight has the potential to enhance state-of-the-art methods that currently rely on IoU scores and localisation losses.

\addtolength{\textheight}{-12cm}   





\section*{ACKNOWLEDGMENT}

Esteban Rivera, as a first author, developed the method, implemented the code and wrote the paper; Loic Stratil contributed with the analysis and wrote the paper, Markus Lienkamp made an essential contribution to the conception of the research project. He revised
the paper critically for important intellectual content. Markus
Lienkamp gave final approval of the version to be published
and agrees to all aspects of the work. The
authors would like to thank the Munich Institute of Robotics
and Machine Intelligence (MIRMI) for their support.


\bibliographystyle{IEEEtran}
\bibliography{egbib}

\end{document}